%% file: 0536.tex
\documentclass[runningheads]{llncs_camera_ready}

\usepackage{times}
\usepackage{cite}
\usepackage[colorlinks=true, citecolor=blue, linkcolor=blue]{hyperref}
\usepackage{epsfig}
\usepackage{graphicx}
\usepackage{subfig}
\usepackage{caption}
\usepackage{multirow}
\usepackage{xspace}

\usepackage{amsmath}
\usepackage{amssymb}

\usepackage{algorithm}
\usepackage[noend]{algpseudocode}

\usepackage{lipsum}
\usepackage{xcolor}
\usepackage[width=122mm,left=12mm,paperwidth=146mm,height=193mm,top=12mm,paperheight=217mm]{geometry}

\makeatletter
\DeclareRobustCommand\onedot{\futurelet\@let@token\@onedot}
\def\@onedot{\ifx\@let@token.\else.\null\fi\xspace}
\def\etal{et al\onedot}
\def\etc{etc\onedot}
\def\ie{i.e\onedot}
\def\eg{e.g\onedot}

\def\vs{vs\onedot}

\newcommand{\textcomm}[1]{``#1''\xspace} 
\definecolor{redcol}{rgb}{1, 0, 0}
\definecolor{bluecol}{rgb}{0, 0, 1}

\DeclareMathOperator*{\argmax}{arg\,max}

\newcommand\blfootnote[1]{%
    \begingroup
    \renewcommand\thefootnote{}\footnote{#1}%
    \addtocounter{footnote}{-1}%
    \endgroup
}

\begin{document}
\pagestyle{headings}
\mainmatter

\title{Zero-Shot Object Detection}

\titlerunning{Zero-Shot Object Detection}
\authorrunning{Bansal \etal}

\author{Ankan Bansal$^*$\inst{1} \and Karan Sikka\inst{2} \and Gaurav
Sharma\inst{3} \and Rama Chellappa\inst{1} \and Ajay Divakaran\inst{2}}
\institute{University of Maryland, College Park, MD \and SRI International, Princeton, NJ \and NEC
Labs America, Cupertino, CA}

\maketitle

\input{abstract}
\input{intro}
\input{related}
\input{approach}
\input{experiments}
\input{conclusion}
\input{acknowledgement}

\clearpage

{\small
    \bibliographystyle{splncs04}
    \bibliography{zero_shot_detn}
}

\end{document}

%% file: abstract.tex
\begin{abstract}

		We introduce and tackle the problem of zero-shot object detection (ZSD), which aims to detect object
		classes which are not observed during training.  We work with a challenging set of object classes, not
		restricting ourselves to similar and/or fine-grained categories as in prior works on zero-shot
		classification. We present a principled approach by first adapting visual-semantic embeddings for ZSD. We
		then discuss the problems associated with selecting a background class and motivate two background-aware
		approaches for learning robust detectors. One of these models uses a fixed background class and the
		other is based on iterative latent assignments.  We also outline the challenge associated with using a
		limited number of training classes and propose a solution based on dense sampling of the semantic label
		space using auxiliary data with a large number of categories.  We propose novel splits of two standard
		detection datasets -- MSCOCO and VisualGenome, and present extensive empirical results in
        both the traditional and generalized zero-shot settings to highlight the
		benefits of the proposed methods. We provide useful insights into the algorithm and conclude by posing
		some open questions to encourage further research. 
        
\end{abstract}

%% file: intro.tex
\section{Introduction}
\label{intro}

\blfootnote{*Most of the work was done when AB was an intern at SRI International. \\
\texttt{\scriptsize \{ankan,rama\}@umiacs.umd.edu,~\{karan.sikka,ajay.divakaran\}@sri.com,~grv@nec-labs.com}
}
Humans can effortlessly make a mental model of an object using only textual description, while machine
recognition systems, until not very long ago, needed to be shown visual examples of every category of
interest. Recently, some
work has been done on \emph{zero-shot} classification using textual descriptions
\cite{xian2017zero}, leveraging progress made on both visual representations
\cite{szegedy2017inception} and semantic text embeddings \cite{pennington2014glove,
mikolov2013efficient, fasttext}. In zero-shot classification, at training time visual
examples are provided for some visual classes but
during testing the model is expected to recognize instances of classes which were not seen, with the constraint that the
new classes are semantically related to the training classes. 

\begin{figure}[t] 
	\begin{center} 
		\includegraphics[width=\linewidth]{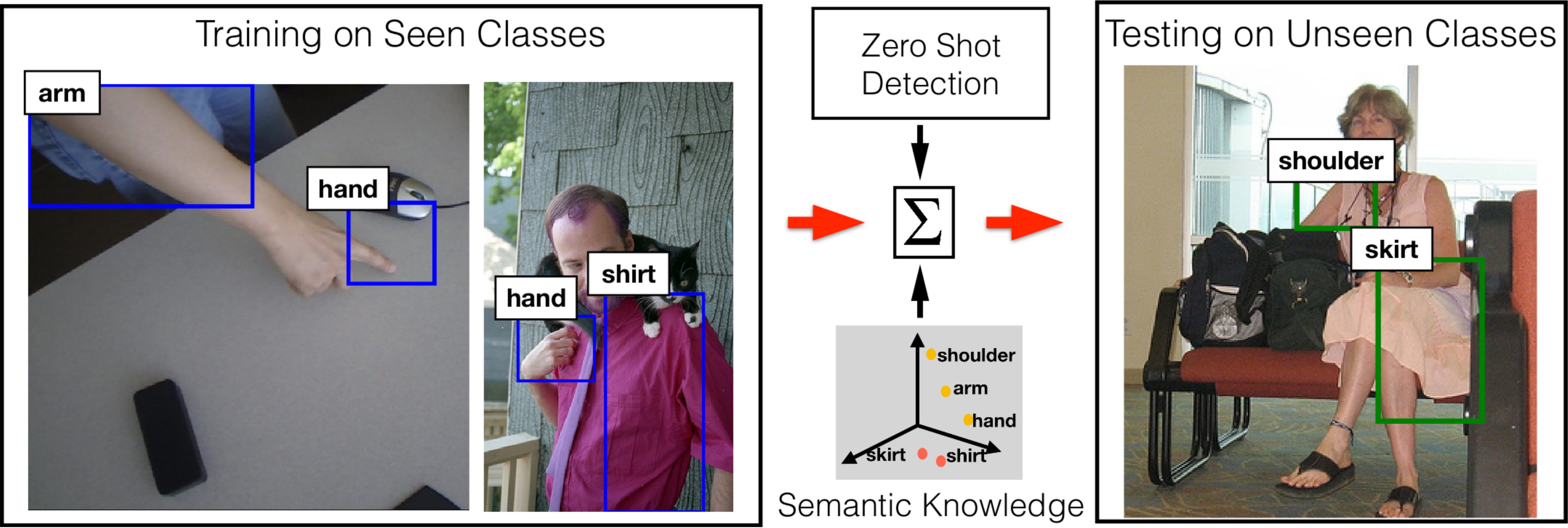}
    \end{center} 
		\caption{We highlight the task of zero-shot object detection where objects \textcomm{arm},
		\textcomm{hand}, and \textcomm{shirt} are observed (seen) during training, but
        \textcomm{skirt},
        and \textcomm{shoulder} are not. These unseen classes are localized by our approach
        that leverages semantic relationships between seen and unseen classes along with the
    proposed zero-shot detection framework. The example has been generated by our model.}
	\label{fig:unseen_example} 
\end{figure}

This problem is solved within the framework of transfer learning \cite{fu2017recent, qi2011towards}, where visual models
for seen classes are transferred to the unknown classes by exploiting semantic relationships between
the two. For example, as shown in figure \ref{fig:unseen_example}, the semantic similarities between classes
\textcomm{hand} and \textcomm{arm} are used to detect an instance of a related (unseen) class \textcomm{shoulder}.
While such a setting has been used for object classification, object detection has remained mostly in the fully
supervised setting as it is much more challenging.  In comparison to object classification, which aims to predict the
class label of an object in an image, object detection aims at predicting bounding box locations for multiple
objects in an image. While classification can rely heavily on contextual cues, \eg airplane co-occurring with clouds,
detection needs to exactly localize the object of interest and can potentially be degraded by contextual correlations
\cite{yu2016role}.  Furthermore, object detection requires learning additional invariance to appearance, occlusion,
viewpoint, aspect ratio \etc in order to precisely delineate a bounding box \cite{hoiem2012diagnosing}. 

In the past few years, several CNN-based object detection methods have been proposed. Early methods 
\cite{girshick2016region, girshick2015fast} started with an object proposal generation step and classified each object
proposal as belonging to a class from a fixed set of categories. More recent methods either generate
proposals inside a CNN \cite{ren2015faster}, or have
implicit regions directly in the image or feature maps \cite{liu2016ssd, redmon2016you}. These methods achieved
significant performance improvements on small datasets which contain tens to a few hundreds of
object categories
\cite{everingham2010pascal, lin2014microsoft}. However, the problem of detecting a large number of classes of
objects has not received sufficient attention. 
This is mainly due to the lack of available annotated data as getting bounding box annotations for thousands of categories
of objects is an expensive process. Scaling supervised detection to the level of classification (tens to
hundreds of thousands of classes) is infeasible
due to prohibitively large annotations costs. Recent works have tried to avoid such annotations, \eg \cite{redmon2016yolo9000}
proposed an object detection method that can detect several thousand object classes by using available (image-level) class
annotations as weak supervision for object detection. 
Zero-shot learning has been shown to be effective in situations where there is a lack of annotated data \cite{fu2012attribute,
fu2015transductive, liu2011recognizing, parikh2012relative, xian2017zero, xu2016heterogeneous, zhang2016zero,
zhang2016zeroa}. 
Most prior works on zero-shot learning have addressed the classification problem \cite{
bendale2016towards, changpinyo2016synthesized, elhoseiny2013write, frome2013devise, jain2014multi,
kodirov2017semantic, lampert2009learning, lampert2014attribute, norouzi2013zero, qiao2017visually,
rahman2017unified, socher2013zero, xian2016latent}, using semantic word-embeddings
\cite{frome2013devise, kodirov2017semantic} or attributes \cite{fu2012attribute,
lampert2014attribute, zhang2016zero, li2014attributes} as a bridge between seen and unseen
classes.

In the present work, we introduce and study the challenging problem of \emph{zero-shot detection}
for 
diverse and general object categories. This problem is difficult owing to the multiple challenges involved
with detection, as well as those with operating in a zero-shot setting. Compared to fully supervised object detection, zero-shot
detection has many differences, notably the following. While in the fully supervised case a background class is added to
better discriminate between objects (\eg car, person) and background (\eg sky, wall, road), the
meaning of \textcomm{background}
is not clear for zero-shot detection, as it could involve both background \textcomm{stuff} as
well as objects from unannotated/unseen classes. This leads to non-trivial practical problems for zero-shot
detection. We propose two ways to address this problem: one using a fixed background class and
the other using a large open vocabulary for differentiating different background regions.
We start with a standard zero-shot classification architecture \cite{fu2017recent} and adapt it for 
zero-shot object detection. This architecture is based on embedding both images and class labels
into a common vector  space. In order to include information from background regions, following supervised object
detection, we first try to associate the background image regions into a single background
class embedding. However, this method can be improved by using a latent assignment based alternating algorithm 
which associates the background boxes to potentially different classes belonging to a
large open vocabulary. Since most object detection benchmark datasets usually
have a few hundred classes, the label space can be sparsely
populated. We show that dense sampling of the class label space by  using additional data improves zero-shot
detection. Along with these two enhancements, we provide qualitative and quantitative results to provide insights into the success
as well as failure cases of the zero-shot detection algorithms, that point us to novel directions towards
solving this challenging problem.

To summarize, the main contributions of this paper are: 
(i) we introduce the problem of zero-shot object detection (ZSD) in real world settings and present
a baseline method for ZSD that follows existing work on zero-shot image classification
using multimodal semantic embeddings and fully supervised object detection; (ii) we discuss some
challenges associated with incorporating information from
background regions and propose two methods for training background-aware detectors; (iii) we examine the problem
with sparse sampling of classes during training and propose a solution which densely
samples training classes using
additional data; and (iv) we provide extensive experimental and ablation studies in traditional and
generalized zero-shot settings to
highlight the benefits and shortcomings of the proposed methods and provide useful insights which
point to future research directions. 

%% file: related.tex
\section{Related Work}
\label{sec:related}

\textbf{Word embeddings.}
Word embeddings map words to a continuous vector representation by encoding semantic similarity
between words. Such representations are trained by exploiting co-occurrences in words in large text
corpuses \cite{fasttext, mikolov2013efficient, word2vec, pennington2014glove}. These word vectors
perform well on tasks such as measuring semantic and syntactic similarities between words. In this
work we use the word embeddings as the common vector space for both images and class labels and thus 
enable detection of objects from unseen categories.
\newline
\textbf{Zero-shot image classification.}
Previous methods for tackling zero-shot classification used attributes, like shape, color, pose or geographical
information as additional sources of information \cite{ferrari2009pose,lampert2009learning,lampert2014attribute}. More
recent approaches have used multimodal embeddings to learn a compatibility function between an image vector and class
label embeddings \cite{akata2015evaluation, akata2013label}.  In \cite{xian2016latent}, the authors augment the bilinear
compatibility model by adding latent variables. The deep visual-semantic embedding model \cite{frome2013devise} used
labeled image data and semantic information from unannotated text data to classify previously unseen image categories.
We follow a similar methodology of using labeled object bounding boxes and semantic information in
the form of unsupervised word embeddings to detect novel object categories. For a more compehensive overview of zero-shot
classification, we refer the reader to the detailed survey by Fu \etal
\cite{fu2017recent}. 
\newline
\textbf{Object detection.}
Early object detection approaches
involved getting object proposals for each image and classifying those object proposals using an
image classification CNN \cite{girshick2016region,girshick2015fast,ren2015faster,xu2018}. More recent
approaches use a single pass through a deep convolution network without the need for object region
proposals \cite{liu2016ssd,redmon2016you}. Recently, Redmon \etal \cite{redmon2016yolo9000}
introduced an object
detector which can scale upto $9000$ object categories using both bounding box and image-level
annotations. 
Unlike this setting, we work in a more challenging setting and do not observe any 
labels for the test object classes during training. We build our detection framework on an
approach similar to the proposal-based approaches mentioned above.
\newline
\textbf{Multi-modal learning.}
Using multiple modalities as additional sources of information has been shown to improve performance
on several computer vision and machine learning tasks. These methods can be used for cross-modal
retrieval tasks \cite{faghri2017vse}, or for transferring classifiers between modalities. Recently, \cite{seehearread}
used images, text, and sound for generating deep discriminative representations which are shared
across the three modalities. Similarly, \cite{bimodal} used images and text descriptions for better
natural language based visual entity localization. In \cite{tanmay}, the authors used a shared
vision and language representation space to obtain image-region and word descriptors that can be shared
across multiple vision and language domains. Our work also uses multi-modal learning for building a robust object detector for
unseen classes. 
Another related work is by Li \etal \cite{li2014attributes}, which learns
object-specific attributes to classify, segment, and predict novel objects. The problem proposed
here differs considerably from this in detecting a large set of objects in unconstrained settings
and does not rely on using attributes.
\newline
\textbf{Comparison with recent works on ZSD}: After completion of this work, we found two parallel works by
Zhu \etal \cite{zhu2018zero} and Rahman \etal \cite{rahman2018zero} that target a similar problem.  Zhu \etal focus on
a different problem of generating object proposals for unseen objects. Rahman \etal
\cite{rahman2018zero} propose a loss formulation that combines max-margin learning and a semantic clustering loss. Their
aim is to separate individual classes and reduce the noise in semantic vectors. A key difference between our work and Rahman \etal is the choice of
evaluation datasets. Rahman \etal use the ILSVRC-2017 detection dataset \cite{imagenet} for training and evaluation.
This dataset is more constrained in comparison to the ones used in our work (MSCOCO and
VisualGenome) because it contains only about one object per image on an average. 
We would also like to note that due to a relatively simpler test setting, Rahman
\etal does not consider the corrruption of the background class by unseen classes as done in this work and by Zhu
\etal. 

%% file: approach.tex
\section{Approach}
\label{sec:approach}

We first outline our baseline zero-shot detection
framework that adapts prior work on zero-shot learning for the current task. Since this approach does not consider
the diversity of the background objects during training, we then present an approach for training a background-aware detector
with a fixed background class. We highlight some possible limitations of this approach and propose a latent
assignment based background-aware model. Finally, we describe our method for densely sampling labels using additional data,
which improves generalization.

\subsection{Baseline Zero-Shot Detection (ZSD)}
\label{baseline:zsd}

We denote the set of all classes as $\mathcal{C} = \mathcal{S} \cup \mathcal{U} \cup \mathcal{O}$,
where $\mathcal{S}$ denotes the set of seen (train) classes, $\mathcal{U}$ the set of unseen (test) classes, and
$\mathcal{O}$ the set of classes that are neither part of seen or unseen classes. 
 Note that our methods do not require a pre-defined test set. We fix the unseen classes here just
 for quantitative evaluation. We work in a
zero-shot setting for object detection where, during training we are provided with labeled bounding
boxes that belong to the seen classes only, while during testing we detect objects from unseen
classes.  We denote an image as $I \in \mathbb{R}^{M \times N \times 3}$, provided bounding boxes
as $b_{i} \in \mathbb{N}^{4}$, and their associated labels as $y_i \in \mathcal{S}$. We extract deep
features from a given bounding box obtained from an arbitrary region proposal method. We denote
extracted deep features for each box $b_{i}$ as $\phi(b_{i}) \in \mathbb{R}^{D_1}$.  
We use semantic embeddings to capture the relationships between
seen and unseen classes and thus transfer a model trained on the seen classes to the unseen 
classes as described later. 
We denote the semantic embeddings for different class labels as $w_{j} \in \mathbb{R}^{D_2}$, which
can be obtained from pre-trained word embedding models such as Glove \cite{pennington2014glove}
or fastText \cite{fasttext}. 
Our approach is based on visual-semantic embeddings where both image and text
features are embedded in the same metric space \cite{frome2013devise, socher2013zero}. We project
features from the bounding box to the semantic embedding space itself via a linear projection, 
\begin{align}
	\psi_{i} = W_{p} \phi(b_i)
\end{align}
where, $W_p \in \mathbb{R}^{D_2 \times D_1}$ is a projection matrix and $\psi_{i}$ is the projected
feature. We use the common embedding space to compute a similarity measure between a projected bounding box
feature $\psi_{i}$ and a class embedding $w_j$ for class label $y_j$ as the cosine similarity
$S_{ij}$ between the two vectors. We train the projection by using a max-margin 
loss which enforces the constraint that the matching score of a bounding box with its true class
should be higher than that with other classes. We define loss for a training sample $b_i$ with class label $y_i$ 
as, 
\begin{align}
    \mathcal{L}(b_i, y_i, \theta) = \sum_{j \in \mathcal{S}, j \neq i} \max(0, m - S_{ii} + S_{ij})
\end{align}
where $\theta$ refers to the parameters of the deep CNN and the projection matrix, and $m$ is the
margin. We also add an additional reconstruction loss to $\mathcal{L}$, as suggested by Kodirov \etal
\cite{kodirov2017semantic}, to regularize the semantic embeddings. In particular, we use the projected box 
features to reconstruct the original deep features and calculate the reconstruction loss as the
squared $L2$-distance between the reconstructed feature and the original deep feature.
During test we predict the label ($\hat{y}_i$) for a bounding box ($b_{i}$) by finding its nearest
class based on the similarity scores with different class embeddings, \ie
\begin{align}
    \hat{y}_i = \argmax_{j \in \mathcal{U}} S_{ij}
	\label{eq:assign}
\end{align}

It is common for object detection approaches to include a background class to learn a robust
detector that can effectively discriminate between foreground objects and background
objects. This helps in eliminating bounding box proposals which clearly do not contain any object
of interest. We refer to these models as background-aware
detectors. However, selecting a background for zero-shot detection is a non-trivial problem as we do
not know if a given background box includes background \textcomm{stuff} in the classical sense \eg
sky, ground \etc or an instance of an unseen object class. We thus train our first (baseline) model
only on bounding boxes that contain seen classes. 

\subsection{Background-Aware Zero-Shot Detection}
\label{sec:bckaware}
While background boxes usually lead to improvements in detection performance for current object
detection methods, for ZSD to decide which background bounding boxes to use is not
straight-forward. We outline two approaches for extending the baseline ZSD model by
incorporating information from background boxes during training.
\newline
\textbf{Statically Assigned Background (SB) Based Zero-Shot Detection.}
Our first background-aware model follows as a natural extension of using a fixed background class in
standard object detectors to our embedding framework.  
We accomplish this by adding a fixed vector for the background class in our embedding space.
Such `statically-assigned' background modeling in ZSD, while providing a way to incorporate
background information, has some limitations. First, we are working with the structure imposed by
the semantic text embeddings that represent each class by a vector relative to other semantically
related classes. In such a case it is difficult to learn a projection that can map all the
diverse background appearances, which surely belong to semantically varied classes, to a single
embedding vector representing one monolithic background class.  Second, even if we are able to learn
such a projection function, the model might not work well during testing. It can map any unseen
class to the single vector corresponding to the background,
as it has learned to map everything, which is not from seen classes, to the singleton background class. 
\newline
\textbf{Latent Assignment Based (LAB) Zero-Shot Detection.}
We solve the problems above by spreading the
background boxes over the embedding space by using an Expectation Maximization (EM)-like algorithm.
We do so by assigning multiple (latent) classes to the background objects and thus covering a wider
range of visual concepts. This is reminiscent of semi-supervised learning algorithms
\cite{seeger2000learning}; we have annotated objects for seen classes and unlabeled boxes for the
rest of the image regions. At a higher level we encode the knowledge that a background box does not
belong to the set of seen classes ($\mathcal{S}$), and could potentially belong to a number of
different classes from a large vocabulary set, referred to as background set and denoted as
$\mathcal{O}$. 

We first train a baseline ZSD model on boxes that belong to the seen classes. We then follow an
iterative EM-like training procedure (Algorithm \ref{alg:bootstrapping}), where, in the first of two
alternating steps, we assign labels to some randomly sampled background boxes in the training set
as classes in $\mathcal{O}$ using our trained model with equation \ref{eq:assign}. 
In the second step, we re-train our detection model with the boxes, labeled as above, included. In
the next iteration, we repeat the first step for another part of background boxes and retrain our
model with the new training data.  This proposed approach is also related to open-vocabulary
learning where we are not restricted by a fixed set of classes \cite{jain2014multi,
zhang2016online}, and to latent-variable based classification models \eg \cite{sharma2017epm}.

\begin{algorithm}[t]
	\caption{LAB algorithm}
	\label{alg:bootstrapping}
	\begin{algorithmic}
		\State{Given: \texttt{annoData} (annotated data), \texttt{bgData} (background/unannotated
               data), $\mathcal{C}$ (set of all classes), 
           $\mathcal{S}$ (seen classes), $\mathcal{U}$ (unseen classes), $\mathcal{O}$ (background
       set), \texttt{initModel} (pre-trained network)}
		\State{\texttt{currModel} $\leftarrow$ train(\texttt{initModel}, \texttt{annoData})}
		\For{$i = 1$ to \texttt{niters}}
		\State{\texttt{currBgData} $\leftarrow\phi$}
		\For{$b$ in \texttt{bgData}} 
        \State // distribute background boxes over open vocabulary minus seen classes
		\State{$b_{new} \leftarrow $predict($b$, \texttt{currModel}, $\mathcal{O}$)}
        \State // $\mathcal{O} = \mathcal{C} \setminus (\mathcal{S} \cup \mathcal{U})$
		\State{\texttt{currBgData} $\leftarrow$ \texttt{currBgData} $\cup \{b_{new}\}$}
		\EndFor
		\State{\texttt{currAnnoData} $\leftarrow$ \texttt{annoData} $\cup$  \texttt{currBgData} }
		\State{\texttt{currModel}$\leftarrow$train(\texttt{currModel},\texttt{currAnnoData})}
		\EndFor
    \State return \texttt{currModel}
	\end{algorithmic}
\end{algorithm}

\subsection{Densely Sampled Embedding Space (DSES)}
\label{sec:augmentation}

The ZSD method, described above, relies on learning a common embedding space that aligns
object features with label embeddings.
~A practical problem in
learning such a model with small datasets is that there are only a small number of seen classes,
which results in a sparse sampling of the embedding space during training. This is problematic
particularly for recognizing unseen classes which, by definition, lie in parts of the embedding
space that do not have training examples. As a result the method may not converge towards the right
alignment between visual and text modalities. To alleviate this issue, we propose to augment
the training procedure with additional data from external sources that contain boxes belonging to
classes other than unseen classes, $y_i \in \mathcal{C}-\mathcal{U}$. In other words, we aim to have
a dense sampling of the space of object classes during training to improve the alignment of the
embedding spaces. We show empirically that, because the extra data being used is from diverse 
external sources and is distinct from seen and unseen classes, it improves the
baseline method.

%% file: experiments.tex
\section{Experiments}
\label{sec:experiments}

We first describe the challenging public datasets we use to validate the proposed approaches, and give the
procedure for creating the novel training and test splits\footnote{Visit
\scriptsize \url{http://ankan.umiacs.io/zsd.html}}. We then discuss the implementation details and the evaluation protocol. Thereafter,
we give the empirical performance for different models followed by some ablation studies and
qualitative results to provide insights into the methods. 

\noindent
\textbf{MSCOCO} 
\cite{lin2014microsoft} We use training
images from the 2014 training set and randomly sample images for testing from the validation set. 
\newline
\textbf{VisualGenome} 
(VG) \cite{krishnavisualgenome}
We remove non-visual classes from
the dataset; use images from part-1 of the dataset for training, and randomly sample images from
part-2 for testing.
\newline
\textbf{OpenImages} 
(OI) \cite{openimages} 
We use
this dataset for densely sampling the label space as described in section \ref{sec:augmentation}.
It contains about $1.5$ million images containing $3.7$ million bounding boxes that span $545$
object categories.\\
\newline
\textbf{Procedure for Creating Train and Test Splits}: For dividing the classes into seen (train) and unseen (test)
classes, we use a procedure similar to \cite{anne2016deep}. We begin with word-vector embeddings for all classes and
cluster them into $K$ clusters using cosine similarity between the word-vectors as the metric. We randomly select
$80\%$ classes from each cluster and assign these to the set of seen classes. We assign the remaining $20\%$ classes
from each cluster to the test set. We set the number of clusters to $10$ and $20$ for MSCOCO and VisualGenome
respectively. Out of all the available classes, we consider only those which have a synset associated with them in the
WordNet hierarchy 
\cite{wordnet} and also have a word vector available.
This gives us $48$ training classes and $17$ test classes for MSCOCO and $478$ training classes and $130$ test classes
for VisualGenome. For MSCOCO, to avoid taking unseen categories as background boxes, we remove all images from the
training set which contain any object from unseen categories. However, we can not do this for VG because the large number
of test categories and dense labeling results in most images being eliminated from the training set. After creating the
splits we have $73,774$ training and $6,608$ test images for MSCOCO, and $54,913$ training and $7,788$ test images for VG. 

\subsection{Implementation Details} 
\label{sec:impl_details}
\textbf{Preparing Datasets for Training}: We first obtain bounding box proposals for each image in
the training set. We construct the training datasets by assigning each proposal a class label from
seen classes or the ``background" class  based on its $\textrm{IoU}$ (Intersection over Union) with
a ground truth bounding box.  Since, majority of the proposals belong to background, we only
include a part of the background boxes. Any proposal with $0<\textrm{IoU}<0.2$ with a ground
truth bounding box is included as a background box in the training set. Apart from these, we also
include a few randomly selected background boxes with $\textrm{IoU} = 0$ with any ground truth
bounding boxes. Any proposal with an $\textrm{IoU}>0.5$ with a ground-truth box is assigned to the
class of the ground-truth box. Finally, we get $1.4$ million training boxes for MSCOCO and $5.8$
million training boxes for VG. We use these boxes for training the two background aware models. As
previously mentioned, we only use boxes belonging to seen classes for training the baseline ZSD
model.  In this case, we have $0.67$ million training boxes for MSCOCO and about $2.6$ million
training boxes for VG.  We train our model on these training sets and test them on the test sets as
described above. 
\newline
\textbf{Baseline ZSD Model}: We build our ZSD model on the RCNN framework that first extracts region
proposals, warps them, and then classifies them.  We use the
Edge-Boxes method \cite{zitnick2014edge} with its default parameters for generating region proposals
and then warp them to an image of size $224\times224$. We use the (pre-trained) Inception-ResNet v2
model \cite{szegedy2017inception} as our base CNN for computing deep features. We project image
features from a proposal box to the $300$ dimensional semantic text space by adding a fully-connected
layer on the last layer of the CNN.  
We use the Adam optimizer \cite{kingma2014adam} with a starting learning rate of
$10^{-3}$ for the projection matrix and $10^{-5}$ for the lower layers. The complete network,
including the projection layer, is first pre-trained on the MSCOCO dataset with the test classes
removed for different models and datasets. For each algorithm, we perform end-to-end training while keeping the word embeddings
fixed. The margin for ranking loss was set to $1$ and the reconstruction loss was added to
max-margin loss after multiplying it by a factor
of $10^{-3}$.  We provide
algorithm specific details below. 
\newline
\textbf{Static Background based ZSD}: In this case, we include the background boxes obtained as
described above in the training set. The single background class is assigned a fixed label vector $[1,
\ldots, 0]$ (this fixed background vector was chosen so as to have norm one similar to the other
class embeddings). 
\newline
\textbf{LAB}: We first create a vocabulary ($\mathcal{C}$) which contains all the words for which we
have word-vectors and synsets in the WordNet hierarchy \cite{wordnet}. We then remove any label from
seen and unseen classes from this set. The size of the vocabulary was about $82K$ for VG and about
$180K$ for MSCOCO. In the first iteration, we use our baseline ZSD model to obtain labels from the
vocabulary set for some of the background boxes. We add these boxes with the newly assigned labels
to the training set for the next iteration (see algorithm \ref{alg:bootstrapping}). We fine-tune the model from the previous iteration using
this new training set for about one epoch. During our experiments we iterate over this process five times.
Our starting learning rates were the same as above and we decreased them by a factor of $10$ after
every $2$ iterations.
\newline
\textbf{Dense Sampling of the Semantic Space}: To increase the label density, we use additional data
from OI to augment the training sets for both VG and MSCOCO. We remove all our test classes from OI
and add the boxes from  remaining classes to the training sets. This led to an addition of $238$ classes to VG
and $330$ classes to MSCOCO during training.  This increases the number of training bounding boxes
for VG to $3.3$ million and to $1$ million for MSCOCO. 

\subsection{Evaluation Protocol}
During evaluation we use Edge-Boxes for extracting proposals for each image and select only those
proposals that have a proposal score (given by Edge-Boxes) greater than $0.07$. This threshold was
set based on trade-offs between performance and evaluation time.
We pass these proposals through the base CNN and obtain a score for each test class as
outlined in section \ref{baseline:zsd}. We apply greedy non-maximal suppression 
\cite{girshick2016region} on all the scored boxes for each test class independently and reject boxes
that have an IoU greater than $0.4$ with a higher scoring box. We use recall as the main evaluation
metric for detection instead of the commonly used mean average precision (mAP). This is because,
for large-scale crowd-sourced datasets such as VG, it is often difficult to exhaustively label
bounding box annotations for all instances of an object. Recall has also been used in prior work on
detecting visual relationships \cite{lu2016visual} where it is infeasible to annotate all possible
instances. The traditional mAP metric is sensitive to missing annotations and will count such
detections as false positives. We define Recall@K as the recall when only the top $K$ detections
(based on prediction score) are selected from an image. A predicted bounding box is marked as true
positive only if it has an IoU overlap greater than a certain threshold {$t$} with a ground truth
bounding box and no other higher confidence predicted bounding box has been assigned to the same ground truth box.
Otherwise it is marked as a false positive.  For MSCOCO we also report the mAP since all object
instances in MSCOCO are annotated.

\subsection{Quantitative Results} 
\label{sec:quant}

We present extensive results (Recall@100) for different algorithms on MSCOCO and VG datasets in table \ref{tab:results_unseen} for
three different IoU overlap thresholds. 
We also show the number of seen,
unseen,
and background classes for each case. During our discussion we report Recall@100 at a
threshold of $\textrm{IoU} \geq 0.5$
unless specified otherwise. 

\begin{table}[t]
	\begin{center}     
		\begin{tabular}{|c|c|c|c|c|c|c|c|c|c|c|c|c|c|}
            \multicolumn{1}{c}{}&\multicolumn{1}{c}{} & \multicolumn{6}{c}{\textbf{MSCOCO}} &
            \multicolumn{6}{c}{\textbf{Visual Genome}} \\ \hline 
            ZSD Method & BG- & \multicolumn{3}{|c|}{\#classes} & \multicolumn{3}{|c|}{IoU} &
            \multicolumn{3}{|c|}{\#classes} & \multicolumn{3}{|c|}{IoU}  \\ \cline{3-14}
                                  & aware & $|\mathcal{S}|$ & $|\mathcal{U}|$ & $|\mathcal{O}|$ &
            $0.4$  &  $0.5$ &  $0.6$ & $|\mathcal{S}|$ & $|\mathcal{U}|$ & $|\mathcal{O}|$& $0.4$ & $0.5$ & $0.6$    \\ \hline
            Baseline & & $48$ & $17$ & $0$ & $34.36$ & $22.14~(0.32)$ & $11.31$ & $478$ & $130$ & $0$ & $8.19$ & $5.19$ & $2.63$  \\ \hline
            SB & \checkmark & $48$ & $17$ & $1$ & $34.46$ & $24.39~(0.70)$ & $12.55$ & $478$ & $130$ & $1$ & $6.06$ & $4.09$
			  & $2.43$   \\ \hline
            DSES & & $378$ & $17$ & $0$ & $\mathbf{40.23}$ & $\mathbf{27.19}~(0.54)$ &
			  $\mathbf{13.63}$ & $716$ & $130$ & $0$ & $7.78$ & $4.75$ & $2.34$   \\ \hline
            LAB & \checkmark & $48$ & $17$ & $343$ & $31.86$ & $20.52~(0.27)$ & $9.98$ & $478$ & $130$ &
              $1673$ & $\mathbf{8.43}$ &
			  $\mathbf{5.40}$ & $\mathbf{2.74}$   \\ \hline

		\end{tabular}
	\end{center}
    \caption{$|\mathcal{S}|, |\mathcal{U}|$, and $|\mathcal{O}|$ refer to the number of seen, unseen
    and the average number of active background classes considered during training respectively.
    BG-aware means background-aware representations. This table shows Recall@100
	performance for the proposed zero-shot detection approaches (see section \ref{sec:approach}) on the two datasets at
	different IoU overlap thresholds with the ground-truth boxes. The numbers in parentheses are
    mean average precision (mAP) values for MSCOCO. The
	number of test (unseen) classes for MSCOCO and VisualGenome are $17$ and $130$ respectively.}
	\label{tab:results_unseen}
\end{table}

On the VG dataset the baseline model achieves $5.19\%$ recall and the static background (SB) model achieves a
recall of $4.09\%$. This marked decline in performance is because all the background boxes are being
mapped to a single vector. In VG some of these background boxes might actually belong to the seen
(train) or unseen
(test) categories. This leads to the SB model learning sub-optimal visual embeddings. However, for
MSCOCO we observe that the SB model increases the recall to $24.39\%$ from the $22.14\%$ achieved
by the baseline model. This is because we remove all images that contain any object from unseen
classes from the training set for MSCOCO. This precludes the possibility of having any background
boxes belonging to the test classes in the training set. As a result, the SB model is not corrupted by
non-background objects and is thus more robust than the baseline.

When we densely sample the embedding space and augment the training classes with additional data,
the recall for MSCOCO increases significantly from $22.14\%$ (for baseline) to $27.19\%$. This shows
that dense sampling is beneficial for predicting unseen classes that lie in sparsely sampled parts
of the embedding space. With dense sampling, the number of train classes in MSCOCO are expanded by a
factor of $7.8$ to $378$. In contrast, VG \emph{a priori} has a large set of seen classes ($478$ versus $48$
in MSCOCO), and the classes expand only by a factor of $1.5$ ($716$) when using DSES. As a result
dense sampling is not able to improve the embedding space obtained by the initial set of categories.  In
such scenarios it might be beneficial to use more sophisticated methods for sampling additional
classes that are not represented well in the training set \cite{gavves2015active, qi2011towards,
lim2011transfer}.

The latent assignment based (LAB) method outperforms the baseline, SB, and DSES on VG. It achieves a
recall of $5.40\%$
compared to $5.19\%$, $4.09\%$ and $4.75\%$ achieved by baseline, SB, and DSES respectively. The consistent improvement
across all IoUs compared to SB, that uses a static background, confirms the benefits of spreading background objects over the
embedding space.  However, LAB gives a lower performance compared to the baseline for MSCOCO ($20.52\%$ by LAB versus
$22.14\%$ by baseline). This is not surprising since the iterations for LAB initialize with a larger set of seen classes for
VG as compared to MSCOCO, resulting in an embedding that covers a wider spectrum of visual space. As a result, LAB is
able to effectively spread the background boxes over a larger set of classes for VG leading to better detections.  
On the other hand, for MSCOCO a sparsely sampled embedding space restricts the coverage of visual concepts leading 
to the background boxes being mapped to a few visual categories. We also see this empirically in the
average number of background classes (set $\mathcal{O}$) assigned to the background boxes during
iterations for LAB, which were $1673$ for VG versus $343$ for MSCOCO. In the remainder of the paper
we focus on LAB method for VG and SB for MSCOCO due to their appropriateness for the respective
datasets.

We observe that the relative class-wise
performance trends are similar to object detection methods, such as Faster
RCNN\footnote{\scriptsize\url{http://cocodataset.org/\#detections-leaderboard}} trained on fully supervised
data. For example, classes such as \textcomm{bus} and \textcomm{elephant} are amongst the best
performing while \textcomm{scissors} and \textcomm{umbrella} rank amongst
the worst in performance.  In addition to these general trends, we also discover some interesting
findings due to the zero-shot nature of the problem. For example, the class \textcomm{cat}, which
generally performs well with standard object detectors, did not perform well with SB.  This results
from having an insufficient number
of semantically related categories for this class in the training set which does not allow the model to effectively capture
the appearance of class \textcomm{cat} during testing.  For such cases we find dense sampling to be useful during training.
The class \textcomm{cat} is one of the top performing categories with DSES. Based on such cases we infer
that for ZSD the performance is both a function of appearance characteristics of the
class as well as its relationship to the seen classes. For VG, the best performing classes, such as \textcomm{laptop},
\textcomm{car}, \textcomm{building}, \textcomm{chair}, seem to have well defined appearance characteristics compared to
bad performing classes, such as \textcomm{gravel}, \textcomm{vent}, \textcomm{garden}, which seem to be more of
\textcomm{stuff} than \textcomm{things}. We also observe that the model is unable to capture any true positive
for the class \textcomm{zebra} and is instead detecting instances of \textcomm{zebra} as either \textcomm{cattle} or
\textcomm{horse}. This is because the model associates a \textcomm{zebra} with a \textcomm{giraffe}, which is
close in the semantic space. The model is able to adapt the detector for the class \textcomm{giraffe} to the
class \textcomm{zebra} but fails to infer additional knowledge needed for a successful detector that a zebra differs
from a giraffe in having white stripes, lower height, and has a body structure similar to a horse. 
Finally, we also
observe that compared to the baseline, LAB achieves similar or better performance on $104$ of $130$ classes on VG.
While for MSCOCO, SB and DSES achieve better or similar performance on $12$ and $13$ classes respectively out of $17$
classes, highlighting the advantages of the proposed models.

\subsection{Generalized Zero-Shot Detection (GZSD)}
The generalized zero-shot learning setting is more realistic than the previously discussed zero-shot
setting \cite{xian2017zero} because both seen
and unseen classes are present during evaluation. This is more challenging than ZSD because it
removes the prior knowledge that the objects at test time belong to unseen classes only.
We use a simple novelty detection step which does not
need extra supervision. Given a test bounding box, $b_i$, we first find the 
most probable train and test classes (see \eqref{eq:assign}) ($\hat{y}^s_i$ and $\hat{y}^u_i$ respectively) and the
corresponding similarity scores ($s_i$ and $u_i$). As the novelty detection step, we check if $u_i$
is greater than some threshold $n_t$. We assign the given bounding box to class $\hat{y}^u_i$ if
$u_i \geq n_t$, otherwise to $\hat{y}^s_i$. For MSCOCO, DSES gives the best performance in the
GZSD setting too. At $n_t = 0.2$, DSES achieves a Recall@100 of $15.02\%$ for seen classes and
$15.32\%$ for unseen classes (harmonic mean (HM) $15.17\%$ \cite{xian2017zero}) at $IoU \geq 0.5$
compared to $14.54\%$ and $10.57\%$ (HM $12.24\%$) for the LAB model and $16.93\%$
and $8.91\%$ (HM $11.67\%$) for baseline.

\subsection{Ablation Studies}

We compare results when considering different number, $K$, of high-confidence detections. We define $K=All$ as the scenario where we consider all boxes
returned by the detector with a confidence score greater than the threshold
for evaluation. We compare LAB and the SB models for VG and MSCOCO respectively, with the
corresponding baseline models in table \ref{tab:results_ablation}.  

\begin{table}[t]
	\begin{center}     
        \resizebox{0.63\textwidth}{!}
        {
		\begin{tabular}{|c|c|c|c|c|c|c|}
            \multicolumn{1}{c}{} & \multicolumn{6}{c}{\textbf{MSCOCO}} \\\hline
			& \multicolumn{3}{|c|}{Baseline} & \multicolumn{3}{|c|}{SB} \\ \hline
			K$\downarrow$ \ IoU$\rightarrow$ & $0.3$ & $0.4$ & $0.5$ & $0.3$ & $0.4$ & $0.5$  \\ \hline
            $All$ & $47.91$ & $37.86$ & $24.47~(0.22)$ & $43.79$ &  $35.58$ & $\mathbf{25.12}~(0.64)$  \\ \hline    
            $100$ & $43.62$ & $34.36$ & $22.14~(0.32)$ & $42.22$ &  $\mathbf{34.46}$ &
            $\mathbf{24.39}~(0.70)$  \\ \hline    
            $80$ & $41.69$ & $32.64$ & $21.01~(0.38)$ & $41.47$ &  $\mathbf{33.98}$ &
            $\mathbf{24.01}~(0.72)$  \\ \hline    
            $50$ & $36.19$ & $27.37$ & $17.05~(0.50)$ & $\mathbf{39.82}$ &  $\mathbf{32.6}$ &
            $\mathbf{23.16}~(0.81)$  \\ \hline    
		\end{tabular}
    }
        \hspace{0.3em}
        \resizebox{0.34\textwidth}{!}
        {
		\begin{tabular}{|c|c|c|c|c|c|}
            \multicolumn{6}{c}{\textbf{VisualGenome}} \\\hline
			\multicolumn{3}{|c|}{Baseline} & \multicolumn{3}{|c|}{LAB} \\ \hline
			$0.3$ & $0.4$ & $0.5$ & $0.3$ & $0.4$ & $0.5$  \\ \hline
			$13.88$ & $9.98$ & $6.45$ & $12.75$ &  $9.61$ & $6.22$  \\ \hline    
			$11.34$ & $8.19$ & $5.19$ & $11.20$ &  $\mathbf{8.43}$ & $\mathbf{5.40}$  \\ \hline    
			$10.41$ & $7.55$ & $4.75$ & $\mathbf{10.45}$ &  $\mathbf{7.86}$ & $\mathbf{5.06}$  \\ \hline    
			$7.98$ & $5.79$ & $3.68$ & $\mathbf{8.54}$ &  $\mathbf{6.44}$ & $\mathbf{4.14}$  \\ \hline    
		\end{tabular}
    }
	\end{center}
    \caption{Ablation studies on background-aware approaches for ZSD. We highlight results
    where the performance is higher for background-aware approaches compared to the corresponding
    baseline. For MSCOCO, the values in parentheses are mAP values.}
	\label{tab:results_ablation}
\end{table}

The difference in performance between the cases $K=All$ and $K=100$ is small,
in general, for the background-aware algorithms unlike the baseline. For example, on MSCOCO
the recall for SB falls by an average (across IoUs) of $1.14\%$ points, compared to a fall of
$3.37\%$ for the baseline. This trend continues further down to $K=80$ and $K=50$ with a gradual
decline in performance as $K$ decreases. This shows that the high confidence detections
produced by our model are of high quality.  

We observe that the background-aware models give better quality detections compared to baselines. The
Recall@K for the corresponding background-aware models are better than the baseline at lower $K$ and
higher IoU threshold values for both datasets. This region represents higher quality detections. This shows that incorporating knowledge from background regions is
an important factor for improving detection quality and performance for ZSD. 

\subsection{Qualitative Results}

\begin{figure}[t]
	\begin{center} 
		\includegraphics[width=1\linewidth]{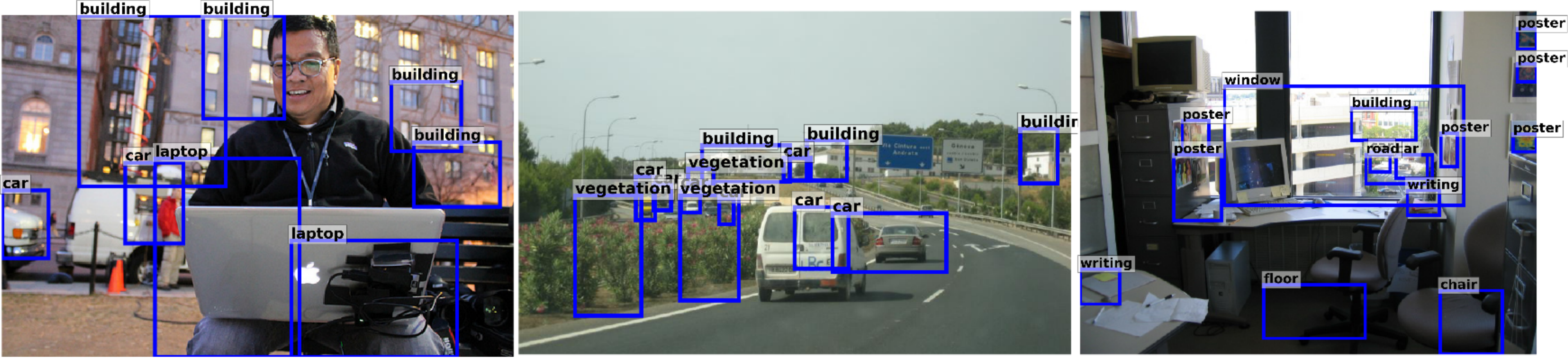} \\
		\includegraphics[width=1\linewidth]{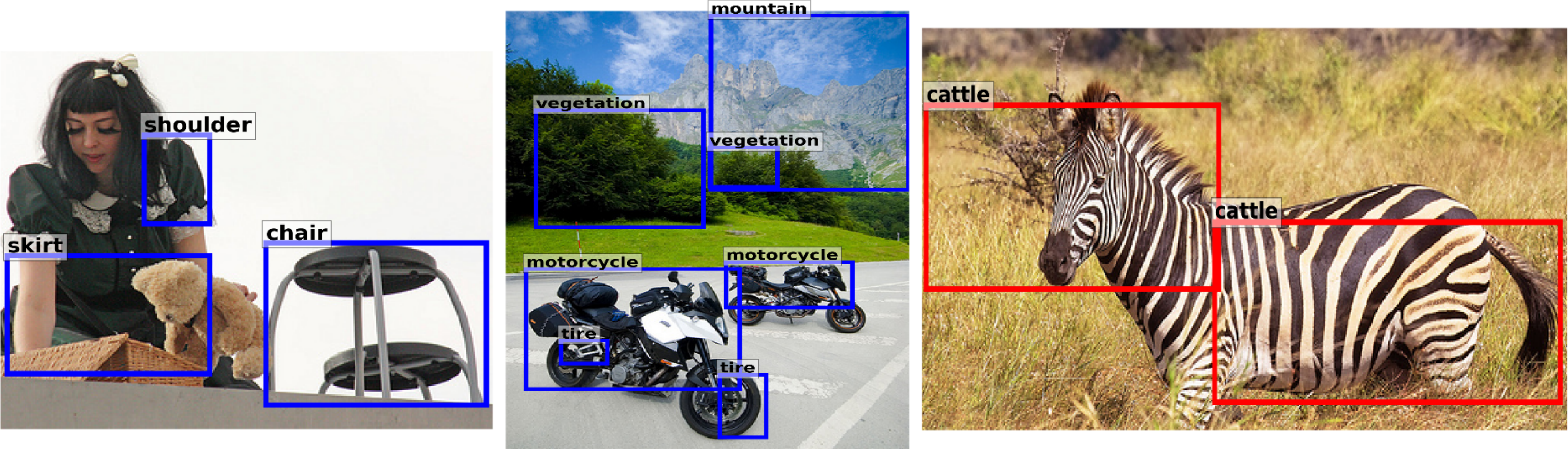} \\
		\includegraphics[width=1\linewidth]{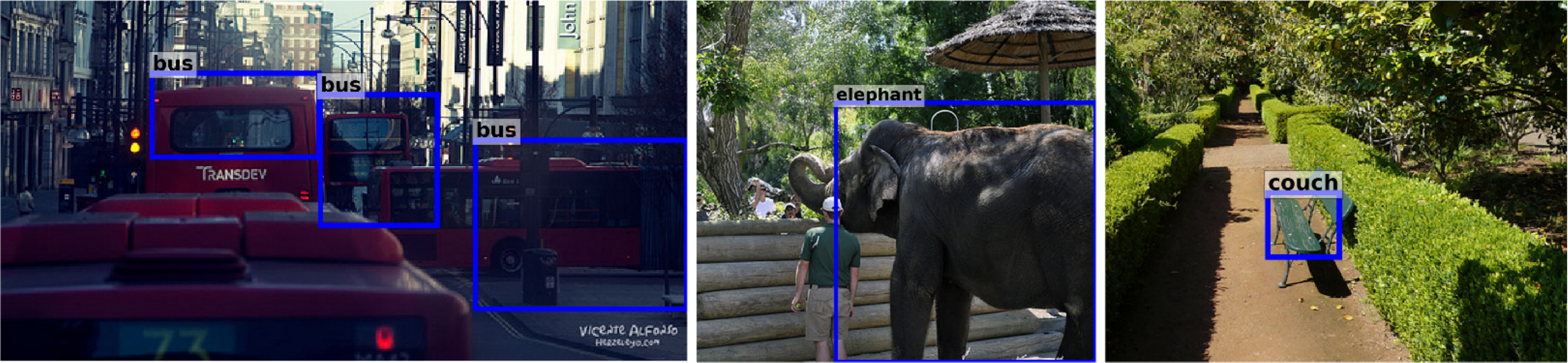} \\
    \end{center} 
	\caption{This figure shows some detections made by the background-aware methods.
	We have used Latent Assignment Based model for VisualGenome (rows $1-2$) and the Static 
    Background model (row $3$) for MSCOCO. Reasonable detections are shown in blue and
two failure cases in red.}
	\label{fig:detection_examples} 
\end{figure}

Figure \ref{fig:detection_examples} shows output detections by the background aware models, \ie LAB
on VisualGenome (first two rows) and SB on MSCOCO (last row). Blue boxes show correct 
detections and red boxes show false positives. These examples confirm that the proposed models are
able to detect unseen classes without observing any samples during training. Further, the models are
able to successfully detect multiple objects in real-world images with background clutter. For
example, in the image taken
in an office ($1^{st}$ row $3^{rd}$ column), the model is able to detect object classes such as
\textcomm{writing}, \textcomm{chair}, \textcomm{cars}. It is also interesting to note that our
approach understands and detects \textcomm{stuff} classes such as \textcomm{vegetation},
and \textcomm{floor}. As discussed in section \ref{sec:quant}, we have shown a failure case
\textcomm{zebra}, that results from having limited information regarding
the fine-grained differences between seen and unseen classes.

%% file: conclusion.tex
\section{Discussion and Conclusion}
\label{sec:conclusion}

We used visual-semantic embeddings for ZSD and addressed the problems
associated with the framework which are specific for ZSD. We proposed two background-aware
approaches; the first one uses a fixed background class while the second iteratively assigns background
boxes to classes in a latent variable framework. We also proposed to improve the sampling density
of the semantic label space using auxiliary data.
We proposed novel splits of two challenging public datasets, MSCOCO and VisualGenome, and
gave extensive quantitative and qualitative results to validate the methods proposed.

Some of the limitations of the presented work, and areas for future work, are as follows. 
It is important to
incorporate some lexical ontology information (\textcomm{is a} and \textcomm{is part of}
relationships) during training and testing for learning models on large vocabularies. Most current
object detection frameworks ignore the hierarchical nature of object classes. For example, a
\textcomm{cat} object should incur a lower loss when predicted as \textcomm{animal} \vs when
predicted as \textcomm{vehicle}. Although a few works have tried to address this issue
\cite{redmon2016you, tanmay}, we believe further work in this direction would be beneficial for
zero-shot detection. We also feel that additional work is needed to generalize bounding-box regression
and hard-negative mining for new objects. 

%% file: acknowledgement.tex
\subsection{Acknowledgements}
This project is sponsored by the Air Force Research Laboratory (AFRL) and Defense Advanced
Research Projects Agency (DARPA) under the contract number USAF/AFMC AFRL FA8750-16-C-0158.
\textbf{Disclaimer}: The views, opinions, and/or findings expressed are those of the author(s) and
should not be interpreted as representing the official views or policies of the Department of
Defense or the U.S. Government.

The work of AB and RC is supported by the Intelligence Advanced Research Projects Activity (IARPA)
    via Department of Interior/Interior Business Center (DOI/IBC) contract number D17PC00345. The
    U.S. Government is authorized to reproduce and distribute reprints for Governmental purposes not
    withstanding any copyright annotation thereon. \textbf{Disclaimer}: The views and conclusions contained
    herein are those of the authors and should not be interpreted as necessarily representing the
    official policies or endorsements, either expressed or implied of IARPA, DOI/IBC or the U.S.
    Government.

We would like to thank the reviewers for their valuable comments and suggestions.